\documentclass[11pt,a4paper]{article}
\usepackage[hyperref]{emnlp2020}
\usepackage{times}
\usepackage{latexsym}

\usepackage[nopar]{lipsum}
\usepackage{graphicx}
\usepackage{algorithm}
\usepackage{algpseudocode}
\usepackage{tabularx}
\usepackage{amsfonts}
\usepackage{hhline}
\usepackage{multirow} 
\usepackage{arydshln}

\newcounter{daggerfootnote}

\newcounter{ddaggerfootnote}

\usepackage{microtype}
\usepackage{amsmath}
\usepackage{enumitem}

\usepackage{microtype}

\aclfinalcopy 

\title{Do We Really Need That Many Parameters In Transformer For Extractive Summarization? Discourse 
Can Help !}

\author{Wen Xiao, Patrick Huber, Giuseppe Carenini\\
  Department of Computer Science \\
  University of British Columbia \\
  Vancouver, BC, Canada, V6T 1Z4 \\
  {\tt \{xiaowen3, huberpat, carenini\}@cs.ubc.ca}}

\date{}

\begin{document}
\maketitle
\begin{abstract}
The multi-head self-attention of popular transformer models is widely used within Natural Language Processing (NLP), including for the task of extractive summarization. With the goal of analyzing and pruning the parameter-heavy self-attention mechanism, there are multiple approaches proposing more parameter-light self-attention alternatives. In this paper, we present a novel parameter-lean self-attention mechanism using discourse priors.
Our new tree self-attention is based on document-level discourse information, extending the recently proposed ``Synthesizer" framework with another lightweight alternative. 
We show empirical results that our tree self-attention approach achieves competitive ROUGE-scores on the task of extractive summarization. When compared to the original single-head transformer model, the tree attention approach reaches similar performance on both, EDU and sentence level, despite the significant reduction of parameters in the attention component. We further significantly outperform the 8-head transformer model on sentence level when applying a more balanced hyper-parameter setting, requiring an order of magnitude less parameters\footnote{Our code can be found here -  \url{http://www.cs.ubc.ca/cs-research/lci/research-groups/natural-language-processing/}}.
\end{abstract}

\section{Introduction}
\label{intro}
\begin{figure}[h!]
    \centering
    \includegraphics[width=0.95\linewidth]{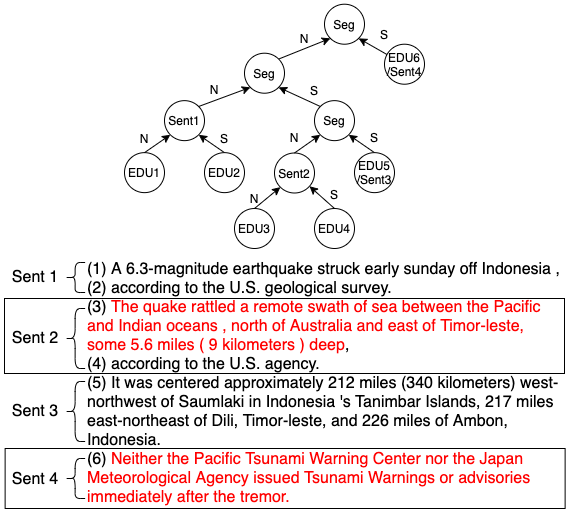}
    \vspace{-3mm}
    \caption{
    News document (4 sentences / 6 EDUs), with both its discourse tree (top) and possible extractive summaries at the sentence/EDU level (extracted sentences and EDUs shown in boxes and red respectively).
    }
    \vspace{-2.5mm}
    \label{fig:example}
\end{figure}
The task of extractive summarization aims to generate summaries for multi-sentential documents by selecting a subset of text units in the source document that most accurately cover the authors communicative goal (as shown in red in Figure \ref{fig:example}). As such, extractive summarization has been a long standing research question with direct practical implications.
The main objective for the task is to determine whether a given text unit in the document is important, generally implied by multiple factors, such as position, stance, semantic meaning and discourse. 


\newcite{Marcu1999} already showed early on that discourse information, as defined in the Rhetorical Structure Theory (RST) \cite{rst}, are a good indicator of 
the importance of a text unit in the given context.
The RST framework, one of the most elaborate and widely used theories of discourse, represents a coherent document (a discourse) as a 
constituency tree. The leaves are thereby called \textit{Elementary Discourse Units} (EDUs), clause-like sentence fragments corresponding to minimal units of content (i.e. propositions). Internal tree nodes, comprising document sub-trees, represent hierarchically compound text spans (or constituents). An additional nuclearity attribute is assigned to each child, representing the importance of the subtree in the local constituent, i.e. the 'Nucleus' child plays a more important role than the 'Satellite' child in the parent's relation. Alternatively, if both children are equally important, both are represented as Nuclei.

While other popular theories of discourse exist (
most notably PDTB \cite{prasadpenn}), RST along with its human-annotated  RST-DT treebank \cite{carlson2002rst}  have been leveraged in the past to  improve extractive summarizations, with either unsupervised \cite{TreeKnapsack,NestedTree}, or supervised  \cite{discourse-aware-extractive} methods.



In this paper, we explore a novel, 
equally important application for discourse information in extractive summarization, 
namely to reduce the number of parameters. Instead of exploiting discourse trees as an additional source of information on top of neural models, we use the information as a prior to reduce the number of parameters of existing neural models. This is critical not only to reduce the risk of over-fitting but also to create smaller models that are easier to interpret and deploy.

Not surprisingly, reducing the number of parameters has become increasingly important in the last years, due to the deep-learning revolution. 
Generally speaking, the objective of reducing neural network parameters involves addressing two central questions: 
\textbf{(1)} What do these models really learn? Such that better priors can be provided and less parameters are required and
\textbf{(2)} Are all the model parameters necessary? To identify which parameters can be safely removed.

 

Recently, researchers have explored these questions especially in the context of transformer models. With respect to what is learned in such models, several experiments reveal 
that the information captured by the multi-head self-attention in the popular BERT model 
(i.e., the learned attention weights) generally align well with syntactic and semantic relations within sentences \cite{analize_lm,kovaleva-etal-2019-revealing}.
Regarding the second question, building on  previous work exploring how to prune large neural models while keeping the performance comparable to the original model \cite{model_prune},  very recently \newcite{synthesizer} has proposed the "Synthesizer" framework, comparing the performance when replacing the dot-product self-attention in the original transformer model with other, less parameterized, attention types. 

Inspired by these two lines of research on transformer-based models, namely the identification of a close connection between learned attention weights and linguistic structures, and the potential for safely reducing attention parameters, we propose a document-level discourse-based attention method for extractive summarization. 
With this new, discourse-inspired approach, we reduce the size of the attention module, the core component of the transformer model, while keeping the model-performance competitive to comparable, fully parameterized models on both EDU and sentence level.



 \vspace{-1mm} 
\section{Related Work}
 \vspace{-2mm} 

\subsection{Attention Methods}
Attention mechanisms have become a widely used component 
of many modern neural NLP models. Originally proposed by \citet{bahdanau2014neural} and \citet{luong2015effective} for 
machine translation, the general idea behind attention is based on the intuition that not all textual units within a sequence contribute equally to the result. Thus, the attention value is introduced to learn how to assess the importance of a unit during training.

In recent years, the role of attention within 
NLP further solidified with researchers exploring new variants,
such as multi-head self-attention, as used in 
transformers
\cite{transformer}. Generally, larger transformer models with more attention-heads (and therefore more parameters) 
achieve better performance for many tasks \cite{transformer}. In the context of explaining the internal workings of neural models, \newcite{kovaleva-etal-2019-revealing} has recently focused on transformer-style models, investigating 
the role of individual attention-heads in the BERT model \cite{bert}. Analyzing the capacity to capture different linguistic information within the self-attention module, they find that information represented across attention-heads is oftentimes redundant, thus showing potential to prune those parameters. 
 
Following these findings, \newcite{fixedencoder_nmt} define a combination of fixed, position-based attention heads and a single learnable dot-product self-attention head. They empirically show that this hybrid approach reduces the spatial complexity of the model, while retaining the original performance. In addition, the hybrid model improves the performance in the low-resource case. Broadening these results, \newcite{synthesizer} further investigate 
the contribution of the self-attention mechanism. In their proposed ``Synthesizer" model, they present a generalized version of the transformer, exploring alternative attention types, generally requiring less parameters, but achieving competitive performances on multiple tasks.

In this paper, instead of pruning the redundant heads of the transformer model empirically or exclusively based on position, we reduce the number of parameters by incorporating linguistic information (i.e. discourse) in the attention computation. We compare our setup for extractive summarization against alternative attention mechanisms, defined in the Synthesizer \cite{synthesizer}.

 \vspace{-1.5mm}  
\subsection{Discourse and Summarization}
 \vspace{-1mm} 
\newcite{Marcu1999} was the first to  explore the application of RST-style discourse to the task of extractive summarization. In particular, he showed that discourse can be used directly to improve summarization, by simply extracting EDUs along the paths with more nuclei as the document summary.

 
Later on, researchers started to 
explore unsupervised methods for discourse-tree-based summarization. \newcite{TreeKnapsack} for example propose a trimming-based method on dependency trees, previously converted from the RST constituency trees, aiming to generate a more coherent summary. 
Based on this idea of trimming the dependency-tree, \citet{NestedTree} propose another method of trimming nested trees, composed into two levels: a document-tree considering the structure of the document and a sentence-tree considering the structure within each sentence. 
 
More recently, further work along this line started to incorporate discourse structures into supervised 
summarization with the goal to 
better leverage the (linguistic) structure of a document. \newcite{xiao-carenini-2019-extractive} and \newcite{discourse_abstractive} thereby use the natural structure of scientific papers (i.e. sections) to improve the inputs of the sequence models, better encoding long documents using a structural prior. They empirically show that such structure effectively improves 
performance. 

  \begin{figure*}[h!]
    \centering
    \includegraphics[width=0.8\linewidth]{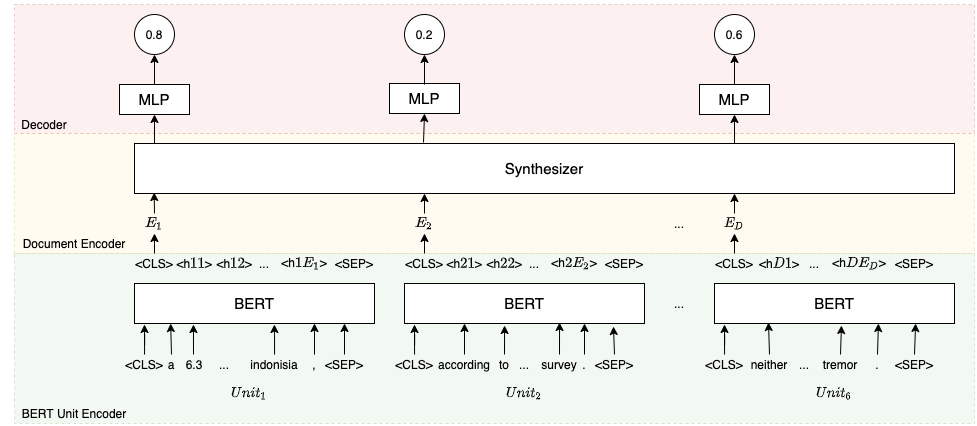}\vspace{-1mm}
    \caption{Structure of the extractive summarization framework containing the Synthesizer module 
    }
    \label{fig:synthesizer_summ}
    \vspace{-0.0mm}
\end{figure*}

Moreover, \newcite{discourse-aware-extractive} propose a graph-based discourse-aware extractive summarization method incorporating the dependency trees converted from RST trees on top of the BERTSUM  model \cite{bertsum} and the document co-reference graph. The results show consistent improvements, implying a close, bidirectional relationship between downstream tasks and discourse parsing. \citet{huber2019predicting, huber2020MEGA} show that sentiment information can be used to infer discourse trees with promising performance. They further mention extractive summarization as another important downstream task with strong potential connections to the document's discourse, motivating the bidirectional use of available information.

This paper employs a rather different objective from aforementioned work combining discourse and summarization. Instead of leveraging additional discourse information to enhance the model performance, we strive to create a summarization model with significantly less parameters, hence being less prone to over-fitting, smaller, and easier to interpret and deploy.
 
 \vspace{-1mm} 
\section{Synthesizer-based Self-Attention Evaluation Framework}
\label{synthesizer}
 \vspace{-2mm} 
Aiming to answer the two guiding questions stated in section \ref{intro}, \newcite{synthesizer} propose a suite of alternative self-attention approaches besides the standard dot-product self-attention, as used in the original transformer model. In their "Synthesizer" framework, they  show that parameter-reduced self-attention mechanisms can achieve competitive performance 
across multiple tasks, including abstractive summarization. While the experiments in the original "Synthesizer" framework are on token level, employing an sequence-to-sequence architecture, we adapt the framework to explore different attention mechanisms on EDU-/sentence-level for the extractive summarization task.

To evaluate the effect of different attention types in our scenario, we apply the general system shown in Figure \ref{fig:synthesizer_summ}, using the pretrained BERT model as our unit encoder. Each unit is thereby represented as the hidden state of the first token in the last BERT layer. Subsequently, we feed the BERT representations into the "Synthesizer" document-encoder \cite{synthesizer} with different attention types and employ a  Multi-Layer Perceptron (MLP) with Sigmoid activation to retrieve a confidence score for each unit, indicating its predicted likelihood to be part of the extractive summary.
\begin{figure*}[h!]
    \centering
    \includegraphics[width=0.9\linewidth]{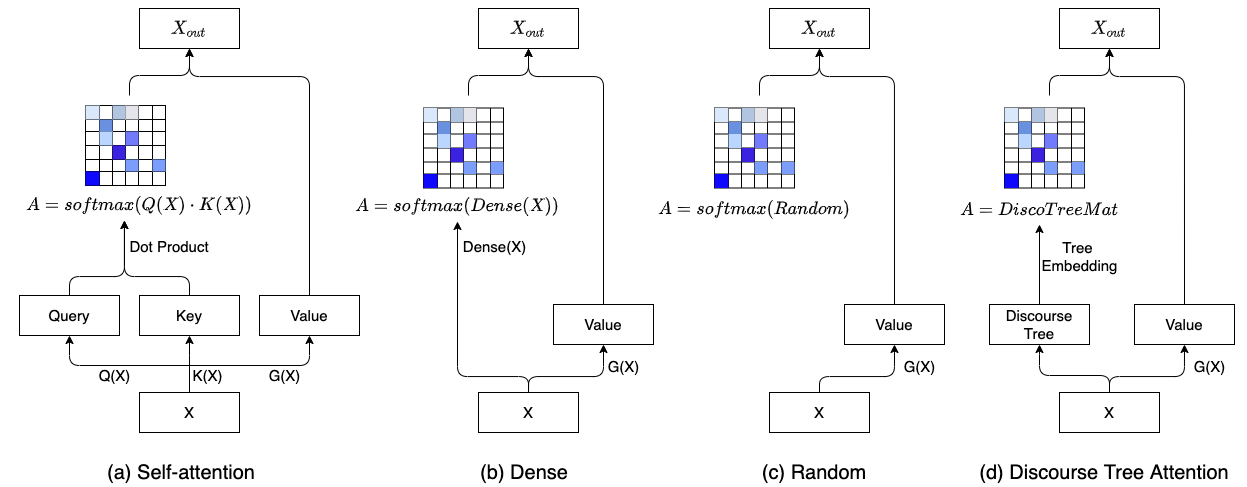}
    \caption{Comparison of attention methods. (a),(b) and (c) taken from \cite{synthesizer}, (d) proposed in this paper.}
    \label{fig:diff_attention}
    \vspace{-2.5mm}
\end{figure*}



The ``Synthesizer" document encoder is essentially a transformer encoder with alternative attention modules, other than the dot-product self-attention. As commonly done, we employ multiple self-attention heads, previously shown to improve the performance of similar models \cite{transformer}. For each attention head, the input is defined as $X \in R^{l\times d}$ where $l$ is the length of the input document (i.e. the number of units), and $d$ represents the hidden dimension of the model. The self-attention matrix is accordingly defined as $A\in R^{l\times l}$, where $A_{ij}$ is the attention-value that unit $i$ pays to unit $j$. We further force the sum of the incoming attentions to each unit (as commonly done) to add up to 1, i.e. $\sum_j A_{ij}=1$. The parameterized function $G$, calculating the \textit{Value}, is multiplied with the attention matrix for the attention output: $X_{out} = A\cdot G(X)$. 
Here, we evaluate the three self-attention methodologies proposed by  \newcite{synthesizer} as our baselines:

\textbf{Dot Product:} As used in the original transformer model, this self-attention calculates a key, a value and a query representation for each textual unit. The attention value is learned as the relationship between the key- and the query-vector defined as $A=softmax(K(X)\cdot Q(X))$

\textbf{Dense:} Instead of using the relationship between units, encoded as keys and values, the dense self-attention $A=softmax(Dense(X))$ is solely learned based on the input unit, where $Dense(\cdot)$ is a two-layer fully connected layer  mapping from $R^{l\times d}$ to $R^{l\times l}$, which can be represented as $Dense(X) =W_1\sigma(W_2X+b_2)+b_1$.\footnote{We use the inner dimension as 512 for all experiments.}

\textbf{Random:} A random attention matrix is generated for each attention-head, shared across all data points, i.e. $A = softmax(R)$. $R$ can thereby be either updated (referred as \textit{Learned Random} in Sec. \ref{experiment}) or fixed (\textit{Fixed Random}) during 
training.
  \vspace{-2mm} 
\section{Discourse Tree Attention}
\label{tree-attention}
 \vspace{-2mm} 
We propose a fourth self-attention candidate: a fixed, discourse-dependent self-attention matrix taking advantage of the strong, tree-structured discourse prior.
(see Figure \ref{fig:diff_attention} for a comparison of all the self-attention methods).
The justification for our new self-attention is two-fold:
\textbf{(1)} RST-style discourse trees represent document-level semantic structures of coherent documents, 
which are important semantic markers for the summarization task 
\textbf{(2)} RST discourse-trees, especially the nuclearity attribute, has been shown to be closely related to the summarization task \cite{Marcu1999, TreeKnapsack, NestedTree}.

To explore a diverse set of RST-style discourse tree attributes, we propose three distinct tree-to-matrix encodings focusing on: the nuclearity-attribute, through a dependency-tree transformation; the plain discourse-structure, derived from the original constituency structure; and a nuclearity-augmented discourse structure, obtained from the constituency representation.
\vspace{-2mm}
\subsection{Dependency-based Nuclearity Attributes (D-Tree)}
\vspace{-1mm}
Inspired by previous work 
using dependency trees to support the summarization task \cite{Marcu1999, TreeKnapsack,discourse-aware-extractive}, 
we first convert the original constituency-tree, obtained with the RST-DT trained discourse parser \cite{wang2017two}, into the respective dependency tree and subsequently generate the final matrix-representation. 

In the first step, we follow the constituency-to-dependency conversion algorithm proposed by \citet{TreeKnapsack} (shown superior for summarization in \citet{hayashi2016empirical}).
While this 
algorithm 
ensures a near-bijective conversion (see \citet{morey2018dependency}), the resulting dependency trees do not necessarily have single-rooted sentence sub-trees.
To account for this, we apply the post-editing method proposed in \citet{hayashi2016empirical}.

\begin{figure}[h!]
    \centering
    \includegraphics[width=\linewidth]{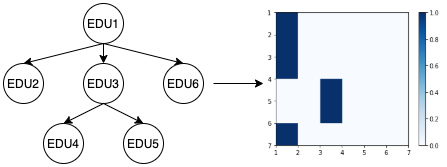}
    \caption{Dependency tree-to-matrix conversion.}
    \label{fig:dep_attention}
    \vspace{-2.5mm}
\end{figure}

To use the newly generated dependency tree in the ``Synthesizer" transformer model, we generate the self-attention matrix from the tree structure by following a standard Graph Theory approach \cite{discourse-aware-extractive}. Head-dependent relations in the tree are represented as binary values (1 indicating a relation, 0 representing no connection) in the self-attention matrix, where each column of the matrix identifies the head and each row represents dependents. The root is considered head and dependent of itself, ensuring all row-sums to be 1. Figure \ref{fig:dep_attention} shows the inferred dependency-tree and the generated self-attention matrix for our running example.
\begin{figure*}[hbt!]
    \centering
    \includegraphics[width=\linewidth]{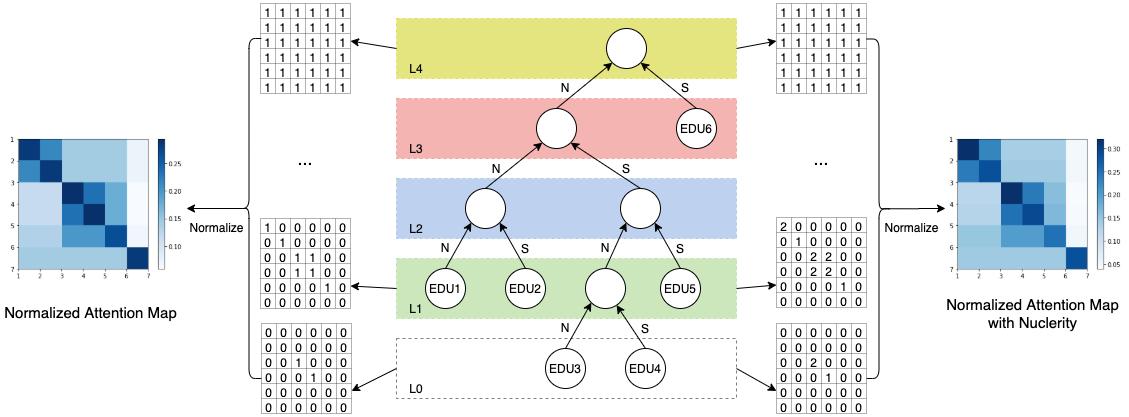}
    \caption{Constituency tree-to-matrix conversion. Left: Structure only, Right: Structure and Nuclearity}
    \label{fig:embed_rstdt}
    \vspace{-2.5mm}
\end{figure*} 
\vspace{-2mm} 
\subsection{Constituency-based Structure Attributes (C-Tree)}
\vspace{-1mm} 
\label{structure_only}

Arguably, there are aspects of the constituency tree-structure that may not be captured adequately by the corresponding dependency-tree. 
These aspects, defining the compositional structure of the document, may contain valuable information for the self-attention. 
In particular, the inter-EDU relationships encoded in the constituency tree can be used to define the relatedness of textual units, implying that the closer the units are in the discourse tree, the more related they are, and the more attention they should pay to each other. Further inspired by the ideas of aggregation \cite{tree-aggregation} and splitting \cite{ordered_neurons}, we define the attention between EDUs based on the depth of the constituency-tree on which they are assigned to the same constituent (Left in Figure \ref{fig:embed_rstdt}).

More specifically, we compute the attention between every two nodes in the self-attention matrix as follows. 
Suppose the height of the constituency-tree is $H$, then for each level $L$ of the tree, there is a binary matrix $M^L\in R^{l\times l}$ with $M^L_{ij}=1$ if EDU $i$ and EDU $j$ are in the same constituent and $M^L_{ij}=0$ otherwise. The final self-attention matrix $A$ is defined as the normalized aggregate matrices of all levels:
$A=normalize(\sum_L M^L)$

The resulting self-attention matrix $A$ is exclusively based on the discourse structure-attribute, without taking the nuclearity into account, representing a rather different approach from the previously  described one based on the dependency-tree. 
 \vspace{-3mm} 
\subsection{Constituency-based Structure and Nuclearity Attributes (C-Tree w/Nuc)}
\vspace{-1mm} 
With the previous sections focusing 
on either exploiting the nuclearity attribute, by converting the RST-style constituency tree into a dependency representation, or the constituency-tree structure itself, we now propose a third, hybrid approach, using both attributes to generate the self-attention matrix. Plausably, the combination could further enhance the quality of the self-attention matrix.
The combined approach is closely related to the structural approach presented in section \ref{structure_only}, but extends the binary self-attention matrix computation to the ternary case. At each level, $M^L_{ij}=2$ if the node rooting the local sub-tree containing EDU $i$ and EDU $j$ is the nucleus in its relation\footnote{The weight of Nucleus and Satellite is set hard-coded, and will be tuned in the future.}, $M^L_{ij}=1$ for the satellite case. Unchanged from section \ref{structure_only}, if EDUs $i$ and $j$ are not sharing a common sub-tree on level $L$, $M^L_{ij}=0$. For example, $M^1_{3:4,3:4}=2$, as the sub-tree containing EDU $3$  \& $4$ is the nucleus in it's relation with the sub-tree containing EDU 5.
 \vspace{-5mm}
\subsection{Sentence-based Discourse Self-Attention}
 \vspace{-1mm}
The natural granularity-level for a discourse-related summarization model is Elementary Discourse Units (EDUs).
Besides using EDUs as our atomic elements, we also explore similar models on sentence-level, the more standard approach in the area of extractive summarization, using a BERT sentence-encoder instead of the previously used BERT EDU-encoder.

To obtain the respective sentence-level self-attention matrix, given the EDU-level self-attention matrix $A^e$ of the three matrix-generation approaches defined above, we define an indicator-matrix $I\in R^{NS\times NE}$. $NS$ and  $NE$ are thereby the number of sentences and EDUs in the document. $I_{ij}=1$ if and only if EDU $j$ belongs to sentence $i$. The sentence-level self-attention matrix $A^s$ is then defined as
\setlength{\belowdisplayskip}{2pt} \setlength{\belowdisplayshortskip}{0pt}
\setlength{\abovedisplayskip}{2pt} \setlength{\abovedisplayshortskip}{0pt}
$$A^s=IA^eI^T$$

Generating the sentence-level self-attention matrices directly from the EDU-level self-attention matrices, instead of the tree-representation itself, avoids the problem of potentially leaky EDUs \cite{joty2015codra}, 
 as sentences with leaky EDUs (having naturally high attention values between them) will continue to be tightly connected.
 \vspace{-2mm}
\section{Experiments}
 \vspace{-1mm}
\label{experiment}
\subsection{Experimental Setup}
\vspace{-1mm}

\textbf{Dataset:} 
We use the popular CNN/DM dataset \cite{nallapati-etal-2016-abstractive}, a standard corpus for 
extractive summarization. 
Key dimensions of the dataset with corresponding statistics are 
in Table \ref{tab:cnndm_stat}.
\begin{table}[h!]
    \centering\resizebox{1\linewidth}{!}{
    \begin{tabular}{c|c|c|c|c}
    \hline
     \#token&\#EDU&\#Sent& \#EDU(O.) &\#Sent(O.)  \\
     \hline
     546&70.2&27.2&6.4&3.1\\
     \hline
    \end{tabular}}
    \caption{Statistics of the CNNDM dataset. O. means the average number of units in the oracle}
    \label{tab:cnndm_stat}
\end{table}
\begin{table*}[h!]
    \centering
    \resizebox{0.8\linewidth}{!}{
    \begin{tabular}{c|c|c|c|c|c|c}
    Model&Rouge-1 &Rouge-2 &Rouge-L& \# Heads&\# Params(attn)&\# Params  \\
    
    \hhline{=======}
    Lead6 &37.99&15.56&34.08&-&-&-\\
    Oracle &62.08&38.20&58.86&-&-&-\\
    DiscoBERT(5 EDUs)&43.77&20.85&40.67&-&-&-\\
    \hline
    \multicolumn{7}{c}{Default Models ($d_k=d_v=d_q=64$, $d_{inner}=3072$)}\\
    \hline
Dot Product(8) &\textbf{41.02}&\textbf{18.78}&\textbf{37.96}&8&3.2M&12.7M\\
\hline
Dot Product(1) &\textbf{40.92}$\ddagger$&\textbf{18.69}$\ddagger$&\textbf{37.85}$\ddagger$&1&0.4M&9.9M\\
Dense &40.70&18.65$\dagger$&37.74$\dagger$&1&1.5M&11.0M\\
Learned Random&40.24&18.28&37.32&1&0.7M&10.3M\\
\hdashline
Fixed Random&40.36&18.35&37.40&1&0.2M&9.7M\\
No attention  &39.89&17.98&36.99&1&0.2M&9.7M\\
D-Tree &40.43&18.32&37.45&1&0.2M&9.7M\\
C-Tree &40.80$\dagger$&18.56&37.74$\dagger$&1&0.2M&9.7M\\
C-Tree w/Nuc &40.76&18.59$\dagger$&37.73&1&0.2M&9.7M\\
\hline
    \multicolumn{7}{c}{Balanced Models ($d_k=d_v=d_q=512$, $d_{inner}=512$)}\\
    \hline
Dot Product(8) &\textbf{40.95}&\textbf{18.52}&\textbf{37.78}&8&25.2M&27M\\
\hline

Dot Product(1) &40.64&18.33&37.54&1&3.2M&4.8M\\
Dense &\textbf{40.87}$\ddagger$&\textbf{18.59}$\ddagger$&\textbf{37.79}$\ddagger$&1&2.9M&4.5M\\
Learned Random&40.32&18.22&37.31&1&2.1M&3.8M\\
\hdashline
Fixed Random&40.18&18.13&37.19&1&1.6M&3.2M\\
No attention  &40.21 &18.17 &37.22 &1&1.6M&3.2M\\
D-Tree &40.29 &18.17 &37.29&1&1.6M&3.2M\\
C-Tree &40.28 &18.13 &37.28 &1&1.6M&3.2M\\
C-Tree w/Nuc &40.70&18.46$\dagger$ $\ddagger$ &37.63 &1&1.6M&3.2M\\ 
\hline
    \end{tabular}}
    \caption{Overall Performance of the models on the EDU level with the number of heads each layer, as well as the number of parameters to train in the attention module and in the whole model. The dashed line splits the models with \textbf{learnt attentions} and with \textbf{fixed attentions}. $\dagger$ indicates that corresponding result is \textbf{NOT} significantly worse than the best result of single-head models with $p<0.01$ with the bootstrap test, and $\ddagger$ indicates that the corresponding result is \textbf{NOT} significantly worse than the result of the 8-head Dot Product with same setting. }
    \label{tab:overall_edu}
    \vspace{-3mm}
\end{table*}
Based on the average number of units selected by the oracle\footnote{The oracle summary contains the units greedily picked according to the ground-truth summary, which is built follow \citet{Kedzie2018}.} on EDU- and sentence-level, we define the summarization task to choose the top 6 EDUs or the highest scoring 3 sentences, depending on the task granularity. Please further note that the original corpus does not contain any EDU-level markers (as presented in Table \ref{tab:cnndm_stat}). The EDU segmentation process employed for EDU-related dataset dimensions is described below.

\textbf{Discourse Augmentation:}
To obtain high-quality discourse representations for the documents in the CNN/DM training corpus we use the pre-trained versions of the top-performing discourse-segmenter \cite{wang2018toward} and -parser \cite{wang2017two}, reaching an F1-score of 94.3\%, 86.0\% (span) and 72.4\% (nuclearity) respectively on the RST-DT dataset.\footnote{We use the publicly available implementation by the original authors at \url{//github.com/yizhongw/StageDP}}
In line with previous work exploring the combination of discourse and summarization, we follow the ``dependency-restriction" strategy proposed in \citet{discourse-aware-extractive} to enhance the coherence and grammatical correctness of the summarization. Such strategy requires that all ancestors of a selected EDU within the same sentence should be recursively added to the final summary.

\textbf{Hyper-Parameters:}
To stay consistent with previous work, we set the dimensions of the attention key($d_k$), value($d_v$) and query vector($d_q$)
to $64$ for each head, and the inner dimension of the position-wise feed-forward layer ($d_{inner}$) to $3072$.
Similar to the synthesizer model \cite{synthesizer}, we only alter the attention part of the transformer model, which contains a small portion of the overall parameters. Additionally, we explore a more balanced, setting, with $d_v=d_k=d_q=512$ and $d_{inner}=512$ for all models.
During training, we use a scheduled learning-rate ($lr=1e-2$) with standard warm-up steps for the Adam optimizer \cite{kingma2014adam}, following the hyper-parameter setting in the original transformer paper \cite{transformer}. 

\textbf{Baseline Models:
}
We compare our new, parameter-reduced Tree Attention approach against a variety of competitive baselines. Based on the standard Dot Product Attention, as used in the original transformer, we explore two settings: A single head and an 8-head Dot Product Attention. Inspired by the "Synthesizer"-framework, we further compare our approach against the Dense and Random Attention computation, as mentioned in Section \ref{synthesizer}. To better show the effect of different attention methods, we use a `No Attention Model' as an additional baseline, in which each input can only attend to itself, i.e. $A=I$. Please note, (1) as our goal is to explore possible parameter reductions, we ensure that all heads contain similar dimensions across models.   
(2) The attention matrices in the ``Fixed Random", ``No Attention" and all three Tree Attention models (D-Tree, C-Tree and C-Tree w/Nuc) are fixed, while they are learned for other models. 
\vspace{-2mm}
\subsection{Results and Analysis}
\vspace{-2mm}
\begin{table*}[h!]
    \centering
    \resizebox{0.8\linewidth}{!}{
    \begin{tabular}{c|c|c|c|c|c|c}
    Model     &Rouge-1 &Rouge-2 &Rouge-L& \# Heads&\# Params(attn)&\# Params  \\
    
    \hhline{=======}
Lead3 &40.30&17.52&36.54&-&-&-\\
    Oracle &56.04&33.10&52.29&-&-&-\\
    BERTSUM(w/Tri-Block) &43.25&20.24&39.63&-&-&118M\\
    \hline
    \multicolumn{7}{c}{Default Models ($d_k=d_v=d_q=64$, $d_{inner}=3072$)}\\
    \hline
Dot Product(8) &\textbf{41.82}&\textbf{19.18}&\textbf{38.18}&8&3.2M&12.7M\\
\hline
Dot Product(1)&\textbf{41.71}$\ddagger$&\textbf{19.08}$\ddagger$&\textbf{38.08}$\ddagger$&1&0.4M&9.9M\\
Dense&41.69$\dagger$&19.07$\ddagger$ $\dagger$&38.08$\ddagger$ $\dagger$&1&1.5M&11.0M\\
Learned Random&41.21&18.86&37.67&1&0.7M&10.3M\\
\hdashline
Fixed Random&41.27&18.91&37.72&1&0.2M&9.7M\\
No attention&40.97&18.64&37.44 &1&0.2M&9.7M\\
D-Tree&41.44&18.87&37.83&1&0.2M&9.7M\\
C-Tree&41.64$\dagger$&19.04$\dagger$&38.03$\dagger$ &1&0.2M&9.7M\\
C-Tree w/Nuc &41.64$\dagger$&19.05$\dagger$&38.03$\dagger$&1&0.2M&9.7M\\
\hline
\multicolumn{7}{c}{Balanced Models ($d_k=d_v=d_q=512$, $d_{inner}=512$)}\\
    \hline
Dot Product(8) &41.45&18.88&37.84&8&25.2M&27M\\
\hline
Dot Product(1)&41.51&18.95&37.94&1&3.2M&4.8M\\
Dense&41.63$\dagger$&19.05$\dagger$&38.01$\dagger$&1&2.9M&4.5M\\
Learned Random&41.26&18.83&37.70&1&2.1M&3.7M\\
\hdashline
Fixed Random&41.17&18.81&37.61&1&1.6M&3.2M\\
No attention &41.25&18.75&37.68 &1&1.6M&3.2M\\
D-Tree&41.31&18.80&37.75&1&1.6M&3.2M\\
C-Tree&\textbf{41.68}&\textbf{19.11}&\textbf{38.12} &1&1.6M&3.2M\\
C-Tree w/Nuc &41.64$\dagger$ &19.02$\dagger$ &38.06$\dagger$ &1&1.6M&3.2M\\
\hline
    \end{tabular}}
    \caption{Overall Performance of the models on the sentence level. $\dagger$ represents that it is \textbf{NOT} significantly worse than the best result of the single-head models with $p<0.01$ with the bootstrap test, and $\ddagger$ indicates that the corresponding result is \textbf{NOT} significantly worse than the result of 8-head Dot Product with same setting ($\ddagger$ for Default Models only).}
    \label{tab:overall_sent}
    \vspace{-3mm}
\end{table*}


We present and discuss three sets of experimental results. First, the natural task for discourse-related extractive summarization on EDU-level. Second, the most common task of extractive summarization on sentence-level and, finally, further experiments regarding the low resource case.

Tables \ref{tab:overall_edu} and \ref{tab:overall_sent} show our experimental results on EDU and Sentence level, respectively. Each row thereby contains the Rouge-1, -2 and -L scores of the model, along with the number of self-attention heads and the amount of trainable parameters in the attention module and in the complete model\footnote{The BERT EDU/sentence encoder is fixed and the parameters therefore not included.}.
For readability, the results in either table are divided into three sub-tables. The first sub-table contains the commonly used Lead-baseline (Lead6 on EDU level and Lead3 on sentence level), 
along with the Oracle, representing the performance upper-bound, and the current state-of-the-art models (DiscoBERT \cite{discourse-aware-extractive} on EDU level, BERTSUM \cite{bertsum} on sentence level). Please note, both SOTA models finetune BERT as a token-based document encoder, to learn additional cross-unit information of tokens. However, this requires additional training resources (as the BERT model itself contains 108M learnable parameters). Furthermore, both SOTA models use 'Trigram-Blocking', which has been shown to be able to greatly improve summarization results \cite{finetune-bert}. The second sub-table shows our experimental results using the default parameter setting, as proposed in the original transformer, and the last sub-table presents the results when using a balanced parameter setting.
Within each sub-table, we further differentiate models by the number of heads, either containing a single attention head or the original 8-head self-attention.
As each document only contains a single discourse tree, there is only one fixed self-attention matrix for each document, making the single-head model equivalent to the multi-head approach.

\noindent\textbf{EDU Level Experiments:} 
are shown in Table \ref{tab:overall_edu}. When comparing the single head models using the default setting (second sub-table), it appears that both, C-Tree and C-Tree w/Nuc achieve competitive performance with the single head Dot Product model, despite the Dot product using twice as many parameters in the attention module ($0.4M$ vs. $0.2M$). This is an important advantage because, even though the non attention related parameters in the complete model outweigh the number of attention parameters in this setting, the attention however resembles the core component of the transformer model, and so saving attention parameters is arguably more critical. In addition, the difference would become large with the increment of the number of heads.
Furthermore, when comparing models with fixed attention or no attention, the effect of the attention module becomes clear, showing superior performance of the C-Tree and C-Tree /w Nuc approaches, indicating that discourse structure can indeed help for the task of extractive summarization. In contrast, the D-Tree inspired self-attention does not perform as well. The drop in performance when using this tree-attention might be caused by the rather strict, binary attention computation, potentially pruning too much valuable discourse information.
Examining the models with learnt attentions, we observe that the Dense model reduces the number of parameters compared to the best performing 8-head Dot Product, however, still contains far more parameters than the single-head Dot Product. Despite the large difference in the number of parameters, the single-head Dot Product Attention performs comparable to the Dense model, suggesting the necessity to synthesize the Dense attention (see \cite{synthesizer}). 
Regarding the Balanced model (bottom sub-table), we put additional emphasis on the attention component, showing trends when using larger attention computation modules. The results in this sub-table suggest that the benefits of our tree attention models are improving over-proportionally for more balanced models, with the C-tree /w Nuc even outperforming the single-head Dot Product and achieving competitive performance to the 8-head Dot Product model, which contains an order of magnitude more parameters in this setting.
\noindent\textbf{Sentence Level Experiments: } 
Our sentence level experiments, presented in Table \ref{tab:overall_sent}, are mostly akin to the EDU level experiments. However, a relative performance improvement of fixed attention models compared to learned attention approaches can be observed, leading to a smaller performance gap on sentence level. We believe that this over-proportional improvement might be due to the position bias on sentence level, which tends to be larger than on EDU level, generally making the sentence level task easier to learn. In line with this trend, we also observe that the difference between the Lead-baseline (40.30/17.53/36.54) and the Oracle (56.04/33.10/52.29) on sentence level is relatively small when compared to the EDU level Lead-baseline(37.99/15.56/34.08) and Oracle(62.08/38.20/58.86). As a result, the fixed tree attention models are statistically equivalent to the learned single-head Dot Product in the default sentence level setting, and significantly outperform the 8-head Dot Product model in the balanced setting.
\noindent\textbf{Low Resource Experiments: } Complementing our previous experiments, showing consistently competitive results of the parameter-sparse models using tree priors, we further explore the robustness of our tree self-attention methods in additional low resource experiments on EDU level. Therefore, we randomly generate 5 small subsets of the training dataset, each containing $1,000$ datapoints, training the same models as shown in Tables \ref{tab:overall_edu} and \ref{tab:overall_sent} on each subset. However, contrasting our initial expectation, the tree-inspired C-Tree w/Nuc model 
only improves the performance on the low-ressource experiments under the balanced setting, with no significant improvements under the default setting. 


   
\noindent\textbf{Overall:} Comparing the results in Tables \ref{tab:overall_edu} and \ref{tab:overall_sent}, it becomes obvious that the sentence level models are consistently better than the EDU level models, despite the opposite trend holding between sentence Oracle and EDU Oracle, as well as the respective SOTA models. One possible reason for this result is that the BERT model is originally trained on the sentences, which might potentially impair the sub-sentential (EDU) representation generation ability of the model.

Furthermore, when comparing the single head and 8 head Dot Product models in both tables and in both settings, we find that the improvement gains of adding additional heads is rather limited, even impairing the performance in the balanced setting on sentence level. We therefore believe that the balance between the performance and the number of parameters is worth of further exploration for the task of extractive summarizarion.


\vspace{-2mm}
\section{Conclusion and Future Work}
\vspace{-1mm}
We extend and adapt the ``Synthesizer" framework for extractive summarization by proposing a new tree self-attention method, based on RST-style consitituency and dependency trees. 
In our experiments, we show that the performance of the tree self-attention is significantly better than other fixed attention models, while being competitive to the single-head standard dot product self-attention in the transformer model on both, the EDU-level and sentence-level extractive summarization task. Furthermore, our tree attention is better than the 8-head dot product in the balanced setting. 
Besides these general results, we further investigate low-resource scenarios, where our parameter-light approaches are assumed to be especially useful. 
However, contrary to this expectation, they do not seem to be more stable and robust than other solutions.
In addition, we also find that the multi-head Dot product model is not always significantly better than the single-head approach. This, combined with the previous finding, suggest that more research is needed on the balance between the number of parameter and the performance of the summarization model.


In the future, we plan to explore ways to also incorporate rhetorical relations into self-attention,  in addition to  discourse structure and nuclearity. Further, we want to replace the hard-coded weight trade-off between Nucleus and Satellite in the C-Tree w/Nuc approach, using instead the confidence score from the discourse parser as the weight.  
Finally, since the current two-level encoder  performs generally worse than a single token-based encoder (e.g. BERTSUM\cite{bertsum}), we intend to explore tree self-attention in combination with the BERTSUM model.



\vspace{-2mm}
\section*{Acknowledgments}
\vspace{-2mm}
We thank 
reviewers and the UBC-NLP group for their insightful comments.
This research was supported by the Language \& Speech Innovation Lab of Cloud BU, Huawei Technologies Co., Ltd.

\bibliography{anthology,emnlp2020}
\bibliographystyle{acl_natbib}

\end{document}